\documentclass[11pt]{article}

\usepackage[preprint]{acl}
\usepackage{amsmath,amssymb,amsthm}
\usepackage{algorithm}
\usepackage{algpseudocode}
\usepackage{times}
\usepackage{latexsym}
\usepackage{comment}
\usepackage[utf8]{inputenc}
\usepackage{xcolor}

\usepackage[T1]{fontenc}

\usepackage[utf8]{inputenc}

\usepackage{microtype}

\usepackage{inconsolata}

\usepackage{graphicx}

\usepackage{booktabs}
\usepackage{tabularx}
\usepackage{multirow}

\newcommand{\errorbox}[1]{%
  \fbox{%
    \begin{minipage}{0.95\linewidth}
    \ttfamily\small #1
    \end{minipage}
  }%
}

\newcommand{\datasource}[1]{%
  \vspace{0.5em}
  {\footnotesize\itshape\color{black!60}Data source: #1.}
}

\usepackage[most]{tcolorbox}

\usepackage{xcolor}
\usepackage{array}
\usepackage{makecell}

\definecolor{mygreen}{RGB}{0,130,0}
\definecolor{myred}{RGB}{180,0,0}

\newtcolorbox{promptbox}[1]{%
  breakable,
  enhanced,
  colback=white,              % 盒子内部白色
  colframe=green!60!black,    % 边框绿色
  arc=6pt,                    % 圆角
  boxrule=0.4pt,              % 边框粗细
  title=#1,                   % 标题文字
  colbacktitle=green!60!black,% 标题背景色（整行）
  coltitle=white,             % 标题文字颜色
  fonttitle=\bfseries\large,  % 标题字体
}
\title{EntroCoT: Enhancing Chain-of-Thought via Adaptive Entropy-Guided Segmentation}

\author{
\textbf{Zihang Li}\textsuperscript{1,2} 
\textbf{Yuhang Wang}\textsuperscript{1} 
\textbf{Yikun Zong}\textsuperscript{1} 
\textbf{Wenhan Yu}\textsuperscript{1}
\textbf{Xiaokun Yuan}\textsuperscript{1} \\
\textbf{Runhan Jiang}\textsuperscript{2}  
\textbf{Zirui Liu}\textsuperscript{1}
\textbf{Tong Yang}\textsuperscript{1$\dagger$}
\textbf{Arthur JIANG}\textsuperscript{2$\dagger$} 
 \\
\textbf{\textsuperscript{1}Peking University}
\textbf{\textsuperscript{2}IQuest Research} \\
\textbf{\textsuperscript{$\dagger$}Corresponding Authors.} %Email: \texttt{\{yangtong@pku.edu.cn, arthursjiang@iquestlab.com}
\vspace{5pt}
}

\begin{document}

\makeatletter
\def\@maketitle{%
  %\newpage
  %\null
  %\vskip 0.5em

  % ---------- logo：左对齐 ----------
  \noindent
  \includegraphics[height=0.9cm]{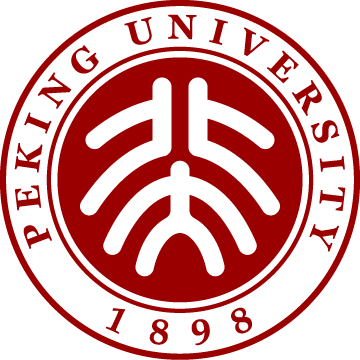}
  %\hspace{0.5mm}
  \includegraphics[height=0.9cm]{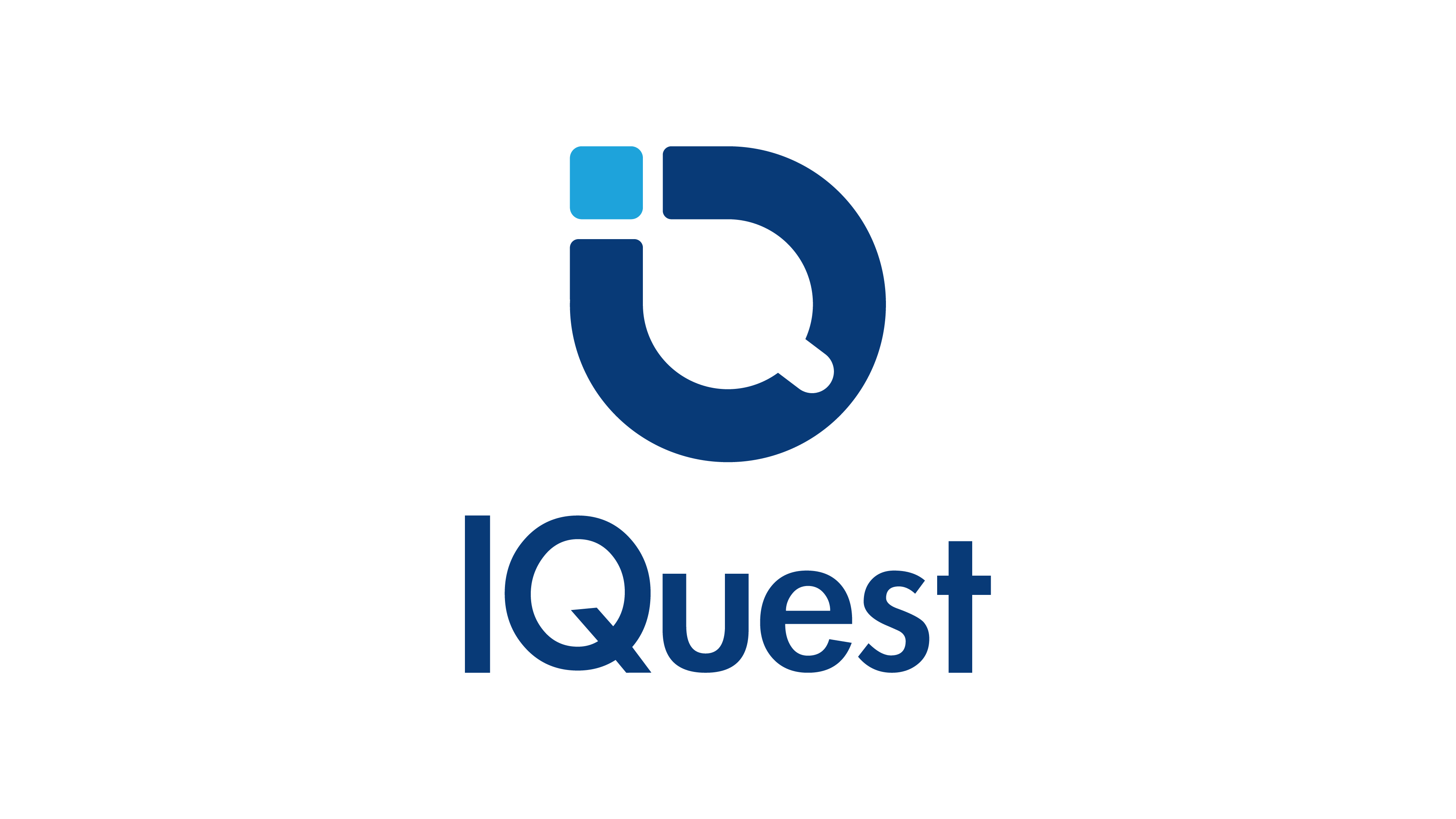}
  \par
  %\vspace{2mm}
  \noindent\rule{\linewidth}{0.5pt}

  \vspace{4mm}

  % ---------- 标题与作者：居中 ----------
  \begin{center}
    {\LARGE\bfseries \@title \par}
    \vskip 1.0em
    {\large
      \lineskip .5em
      \begin{tabular}[t]{c}
        \@author
      \end{tabular}\par}
  \end{center}

  \vskip 1.0em
}
\makeatother

\maketitle 
\begin{abstract}
Chain-of-Thought (CoT) prompting has significantly enhanced the mathematical reasoning capabilities of Large Language Models. 
We find existing fine-tuning datasets frequently suffer from the "answer right but reasoning wrong" probelm, where correct final answers are derived from hallucinated, redundant, or logically invalid intermediate steps.
This paper proposes \textbf{EntroCoT}, a unified framework for automatically identifying and refining low-quality CoT supervision traces. 
EntroCoT first proposes an entropy-based mechanism to segment the reasoning trace into multiple steps at uncertain junctures, and then introduces a Monte Carlo rollout-based mechanism to evaluate the marginal contribution of each step. 
By accurately filtering deceptive reasoning samples, EntroCoT constructs a high-quality dataset where every intermediate step in each reasoning trace facilitates the final answer.
Extensive experiments on mathematical benchmarks demonstrate that fine-tuning on the subset constructed by EntroCoT consistently outperforms the baseslines of full-dataset supervision. 
%Our code is available in "software" appendix.
\end{abstract}

% ------------------ Introduction---------------

\section{Introduction}

Large language models (LLMs) have recently demonstrated strong performance on complex mathematical reasoning tasks, a success largely attributed to the adoption of \emph{chain-of-thought} (CoT) prompting~\cite{wei2022chain}. 
By explicitly decomposing a problem into intermediate reasoning steps, CoT enables models to emulate human-like multi-step reasoning and has become a standard technique for improving accuracy on solving mathematical problem. 
Nowadays, many recent fine-tuning datasets include explicit CoT annotations, either written by humans or synthetically generated by models~\cite{zhao2025promptCoT}.

% The efficacy of CoT-based fine-tuning is increasingly challenged by the presence of logically flawed reasoning trajectories that nevertheless yield correct final answers.
%
% Existing CoT training paradigms largely neglect the logical rigor of intermediate steps, relying instead only on the final answer as the indicator for data quality.
Existing CoT training paradigms often rely on final answers as the sole quality metric, which neglect the logical integrity within CoT traces.
We observe that there are many samples with correct final answer but logically redundant or wrong reasoning steps~\cite{lyu2023faithful}, especially in large-scale synthetic datasets \cite{luo2025deconstructing}. 
For instance, a reasoning trace may erroneously apply the geometric mean formula to a question requiring an arithmetic mean; if the input values are identical, this flawed derivation will coincidentally yield the correct final result.
More examples can be found in Appendix~\ref{app:case_study}.
This ``answer right but reasoning wrong'' issue forces models to mimic logically flawed  patterns and ultimately impairs their ability to solve difficult mathematical tasks \cite{xu2025mind}.
Empirically, we also find that selective training on filtered subsets yields superior performance, indicating many CoT data is counterproductive.
Consequently, establishing a precise definition of CoT trace quality and rectifying logical errors within those traces are critical for effective model tuning \cite{manakul2023selfcheckgpt}.

% , the ground truth of a reasoning trace should be redefined: it should be evaluated not by its surface-level alignment with a template, but by its functional behavioral impact on a model’s ability \cite{manakul2023selfcheckgpt}.

% This work presents a unified framework for the automated identification and correction of misleading CoT supervision.
% % by leveraging a token-level entropy-guided, rollout-based filtering and recovery procedure.
% Our approach is grounded in the empirical observation that increased token-level entropy often signals the emergence of logical inconsistencies and unreliable reasoning \cite{wang2025beyond}. 
% As illustrated in Figure \ref{fig:intro}, high-entropy positions can be treated as forks that could introduce alternative reasoning branches. 
% Therefore, we conceptualize high-entropy tokens as ``decision points of ambiguity'' -- critical junctures where model's reasoning trajectory is most uncertain and multiple logical paths diverge. 
% % ; they mark the specific locations where errors are most likely to be initiated, as well as where logical recovery remains possible. 
% By analyzing the model's behavioral response during rollouts at these pivotal segments, our framework effectively distinguishes between reasoning steps that contribute meaningfully to the final answer and those that are redundant or deceptive, ensuring high-fidelity supervision for model training.

\begin{figure}[t]
    \centering
    \includegraphics[width=0.5\textwidth]{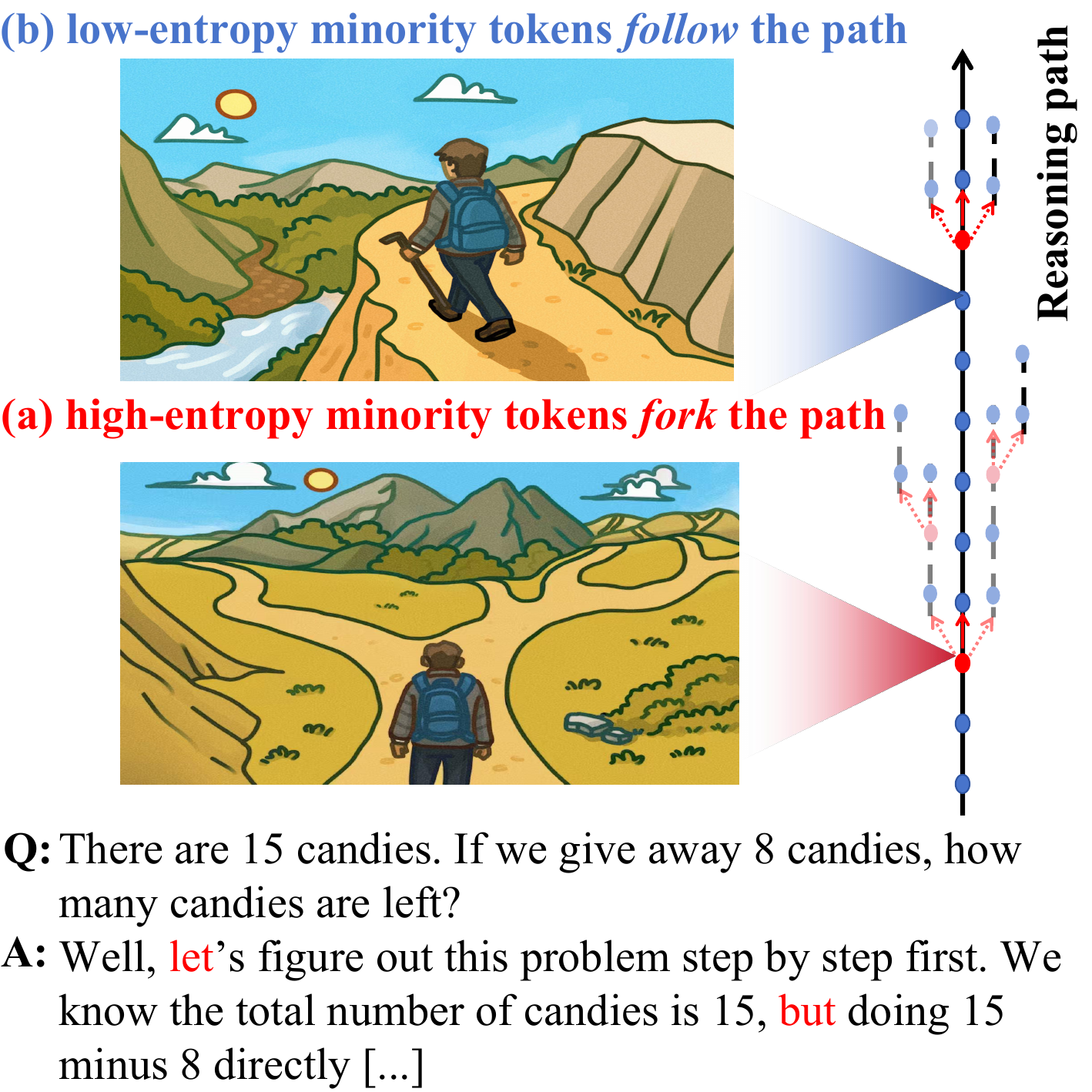}
    \caption{In chain-of-thought reasoning, high-entropy tokens function as forks introducing alternative reasoning branches, whereas low-entropy tokens proceed along the established path.}
    \label{fig:intro}
\end{figure}

To address this issue, we propose a unified framework called \textbf{EntroCoT} for the automated identification and correction of misleading CoT traces.
Our approach is grounded in the observation that high-entropy tokens often signal the emergence of logical inconsistencies \cite{wang2025beyond}, representing critical forks where the model's reasoning becomes uncertain and multiple logical paths diverge.
% Our framework features an entropy-guided segmentation approach designed to adaptively identify and refine deceptive segments within the reasoning trace, which is motivated by the observation that high-entropy tokens often signal the emergence of logical inconsistencies and unreliable reasoning \cite{wang2025beyond}. 
As illustrated in Figure \ref{fig:intro}, we conceptualize high-entropy tokens as critical forks where model's reasoning trace is most uncertain and multiple possible logical paths diverge. 
After partitioning the trace at these pivotal junctures, our framework systematically evaluates the marginal contribution of each segment to the final answer and adaptively filters those deceptive samples. 
Finally, we guarantee that each step in the reliable CoT trace consistently promotes the correct solution, thereby securing a high-fidelity reasoning process.

% This work proposes a unified framework to automatically identify and remove low-quality or misleading CoT supervision. 
% Empirical observations suggest that high-entropy tokens often correlates with the emergence of logical inconsistencies and unreliable deductions~\cite{wang2025beyond}. 
% Motivated by this observation, we introduce an entropy-guided, rollout-based filtering procedure for the reasoning process in data.
% Rather than treating high-entropy positions as merely unreliable, we interpret them as decision points of ambiguity where the model’s output is most uncertain and multiple reasoning paths may be possible. These positions often indicate both the location where an error is most likely conducted and the point where recovery is most possible. As such, we treat them as anchors for exploration, which is shown in Figure~\ref{fig:intro}. By analyzing model behavior around these segments, we can more effectively identify which parts of the reasoning trace contribute meaningfully to the final answer and which may be misleading or redundant.

Guided by this philosophy, EntroCoT incorporates three stages to assess and filter reasoning traces. 
In the first stage, we compute token-level entropy across the generated CoT trace using a strong teacher model, leveraging high-entropy tokens as ``logical anchors'' to partition the trace into discrete segments at critical junctures where the trace is most uncertain.
In the second stage, we introduce a Monte Carlo rollout-based prefix evaluation mechanism to quantify the marginal contribution of each segment to final answer. 
Finally, the filter process begins: should a segment yield a negative impact on performance, the framework automatically filters the corresponding misleading reasoning path(sample). This pipeline is repeated until each sample in the dataset has been filtered, effectively eliminating samples containing deceptive intermediate steps.
 % leading to models that not only answer correctly but reason correctly.
In this way, EntroCoT enables a scalable alignment between training data and high-fidelity reasoning, yielding models that exhibit both conclusion accuracy and logical integrity.

Extensive experiments across various benchmarks, base models, and training datasets demonstrate that EntroCoT consistently improves the average accuracy by 2.71\% for Llama-3.1-8B and 5.17\% for Qwen2.5-Math-1.5B compared to vanilla training on full dataset for Numinamath. 
We also conduct ablation studies to validate the design details of our entropy-guided segmentation.

This paper makes the following contributions.

\begin{itemize}
    \item We identify and formalize a critical bottleneck of existing CoT fine-tuning traces, where the final answer is correct but the intermediate CoT steps are wrong. We demonstrate that such deceptive traces impair model performance. 
    
    \item We propose EntroCoT, a unified framework to assess the quality of CoT data and filter misleading traces. EntroCoT features an entropy-based segmentation mechanism and is compatible to existing fine-tuning workflows. 

    \item We conduct extensive evaluation showing EntroCoT can effectively improve fine-tuning accuracy compared to full-dataset supervision.  
    
    % \textbf{Segmental Filtering Framework}: We introduce a fully automated framework that evaluates CoT quality through a "locate-and-verify" approach. By combining token-level entropy to identify fragile reasoning anchors with segmental rollout analysis, our method assesses the behavioral utility of reasoning steps without requiring human annotations or symbolic solvers.
    
    % \item \textbf{Universal Compatibility and Efficiency}: Our framework is designed as a modular module compatible with various reasoning datasets. We demonstrate that by filtering out low value trajectories, it significantly reduces training compute costs while yielding a more robust model compared to training on the complete, noisy dataset.
    
    % \item \textbf{Empirical Validation of Quality over Volume}: Comprehensive experiments show that fine tuning on only the top $55\%$ of data selected by our pipeline consistently outperforms full-dataset baselines. This finding validates that optimizing the quality of reasoning supervision is fundamentally more impactful for eliciting reasoning capabilities than merely increasing data volume.
\end{itemize}

% By connecting reasoning trace quality directly to task performance, this work offers a scalable way to align training data with desired reasoning behavior, leading to models that not only answer correctly but reason correctly.

% ----------------------------Related Work----------------------

\section{Related Work}

The advancement of Large Language Models (LLMs) has been significantly driven by the Chain-of-Thought (CoT) reasoning paradigm. As models scale, ensuring the quality of these reasoning chains has become paramount for effective model distillation. Our work draws inspiration from and builds upon recent developments in CoT distillation, entropy-based reasoning analysis, and process-level supervision.

\subsection{CoT Distillation and Data Synthesis}
Distilling reasoning capabilities from strong teacher models (e.g., OpenAI o1, GPT-4) to smaller student models is a widely adopted strategy. However, the reliability of the teacher's intermediate reasoning steps remains a challenge. 
Zhao et al. proposed \textit{PromptCoT} \citep{zhao2025promptCoT}, which synthesizes Olympiad-level problems by mimicking the rationale generation process of human experts, highlighting the importance of high-quality reasoning paths for data generation. 
Similarly, Xiang et al. introduced \textit{Meta-CoT} \citep{xiang2025towards}, which models the "meta-reasoning" process to supervise the generation of synthetic data, aiming to emulate System 2 thinking. 
While these methods focus on \textit{generating} new data, our work addresses the challenge of \textit{verifying} and \textit{filtering} existing long-CoT datasets (such as OpenR1), which often contain correct final answers but hallucinated or logically flawed intermediate steps.

\subsection{Entropy-based Reasoning Analysis}
Entropy, as a measure of uncertainty in next-token prediction, has emerged as a powerful tool for analyzing the internal dynamics of LLM reasoning. 
Cheng et al. \citep{cheng2025reasoning} demonstrated that high-entropy tokens often correlate with exploratory behaviors, such as pivotal turning points or self-corrections within a reasoning chain. 
Leveraging this property, Li et al. \citep{li2025compressing} proposed using step entropy to identify and remove redundant, low-information steps, thereby compressing CoT without sacrificing accuracy. 
In parallel, Wang et al. introduced \textit{R1-Compress} \citep{wang2025r1}, which combines chunk compression with search strategies to optimize long reasoning chains.
Different from these works which utilize entropy primarily for \textit{compression} or \textit{exploration}, we propose to use the spatial distribution of high-entropy points (across the beginning, middle, and end of the reasoning process) as a signal for \textit{structural segmentation}. We hypothesize that these high-entropy points represent logical junctions that require verification.

\begin{figure}[H]
    \centering
    \includegraphics[width=0.5\textwidth]{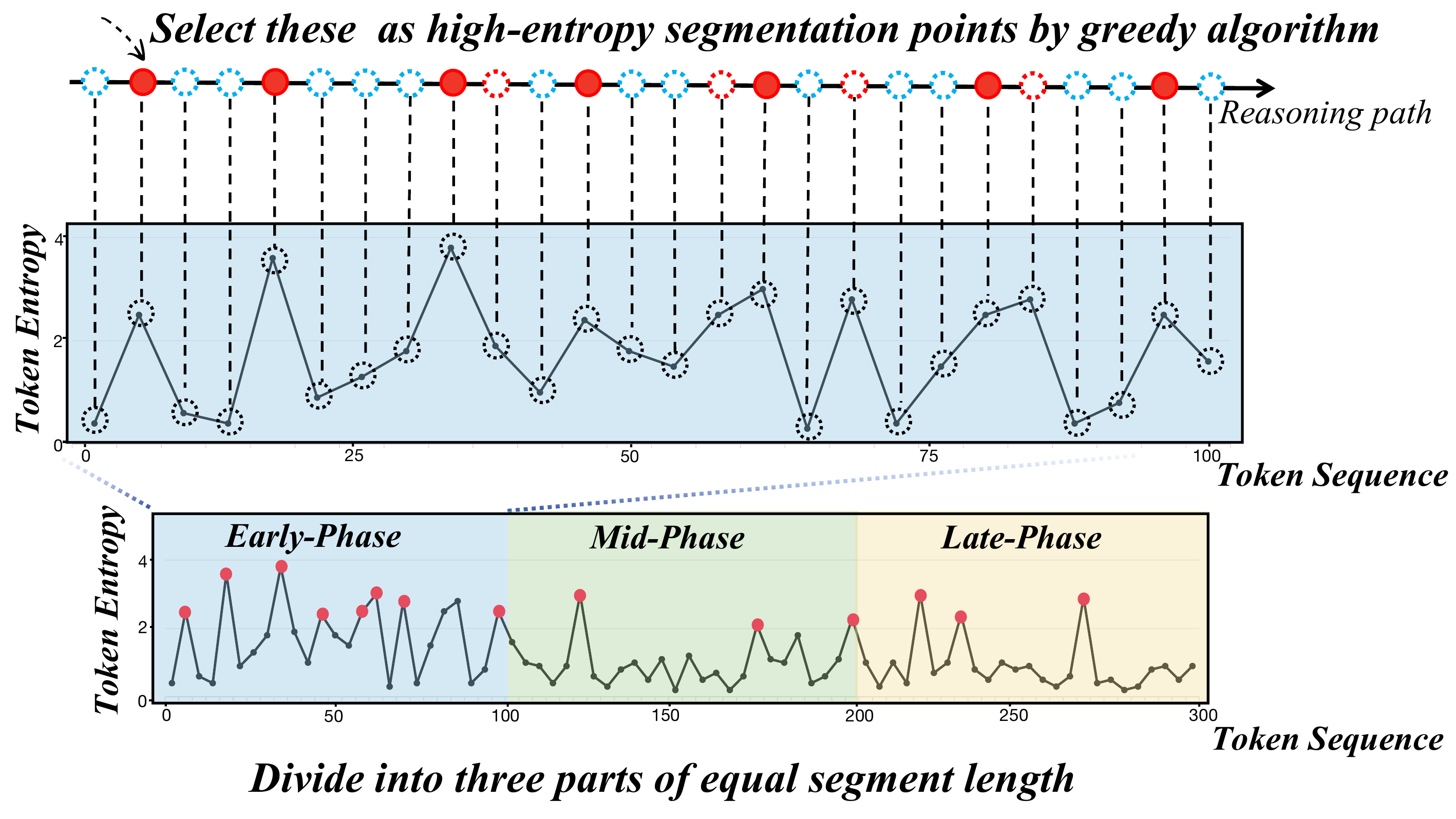} 
    \caption{Token-level entropy is computed along the CoT reasoning path, and the sequence is evenly divided into early, middle, and late segments by token order. Based on the distribution of high-entropy positions within each segment, a greedy algorithm selects spatially dispersed high-entropy positions as segmentation points (red dots) for subsequent construction of multiple prompts.}
    \label{fig:seg}
\end{figure}

\subsection{Structured Process Supervision}
To mitigate the "reasoning gap" where a model arrives at the correct answer through erroneous steps, researchers have turned to fine-grained process supervision. 
Zhang et al. \citep{zhang2025lessons} provided a comprehensive analysis of Process Reward Models (PRMs) in mathematical reasoning, emphasizing the necessity of step-by-step verification. 
Luo et al. proposed \textit{DLCoT} \citep{luo2025deconstructing}, a framework that deconstructs long CoT into segments for simplified optimization and error correction. 
Furthermore, Xu et al. addressed the issue of "Thought Leaps" in \textit{Mind the Gap} \citep{xu2025mind}, proposing methods to bridge disconnected reasoning steps during fine-tuning.
Our approach aligns with the philosophy of process supervision but introduces a novel, computation-efficient verification mechanism. Instead of training a heavy PRM, we utilize a smaller model to "rollout" from entropy-determined breakpoints. By monitoring the non-decreasing trend of accuracy across segments, we can identify reliable reasoning paths and selectively employ strong models to repair fractured logic, thus constructing a high-quality dataset for robust distillation.

% --------------------------Method ---------------------

\section{Method}
\label{sec:method}
While the responses generated by teacher models are typically correct in the final output, they may contain redundant, inconsistent, or locally erroneous reasoning steps. This section introduces an automated method for filtering reasoning data (e.g., Chain-of-Thought, CoT). The core objective of this approach is to identify responses that exhibit \emph{stable positive guidance} for student models at the intermediate reasoning level.

The process is structured into two main stages:
(1) Token-level entropy-based response segmentation;
(2) Prefix evaluation using Monte Carlo rollouts.
The overall algorithmic flow is depicted in Algorithm~\ref{alg:CoT_filtering}.

\subsection{Entropy-Guided CoT Segmentation}
\label{subsec:segmentation}
Given a sample $(x, y, \mathcal{T})$, where $x$ represents the input question, $y$ is the final answer, and $\mathcal{T}=(t_1,\dots,t_L)$ is the complete reasoning sequence generated by the teacher model, the following procedure is applied.

First, we compute the token-level entropy for each token in the sequence $\mathcal{T}$, using the teacher model $\mathcal{M}_T$ conditioned on the input $x$:
\[
H_i = - \sum_{v \in \mathcal{V}} p_{\mathcal{M}_T}(v \mid x, t_{<i}) \log p_{\mathcal{M}_T}(v \mid x, t_{<i}),
\]
where $\mathcal{V}$ denotes the model's vocabulary. The $K$ highest entropy positions are identified, forming the index set $\mathcal{H} = \{h_1, \dots, h_K\}$.

We use Figure~\ref{fig:seg} to illustrate the specific CoT segmentation procedure.
We first divide the sequence $\mathcal{T}$ is into three regions: the beginning, middle, and end. Then we count the number of high-entropy positions in each region, yielding the values $(r_1, r_2, r_3)$, where $r_1 + r_2 + r_3 = K$.

Given the target number of segments $N$, the number of segments for each region is determined based on the proportion of high-entropy positions in each region:
\[
s_i = \left\lfloor N \cdot \frac{r_i}{r_1 + r_2 + r_3} \right\rfloor, \quad i \in \{1,2,3\},
\]
with a normalization step to ensure that $s_1 + s_2 + s_3 = N$. This allocation strategy ensures that regions with higher uncertainty are divided into more reasoning sub-segments.

\begin{figure*}[t]
    \centering
    \includegraphics[width=1\textwidth]{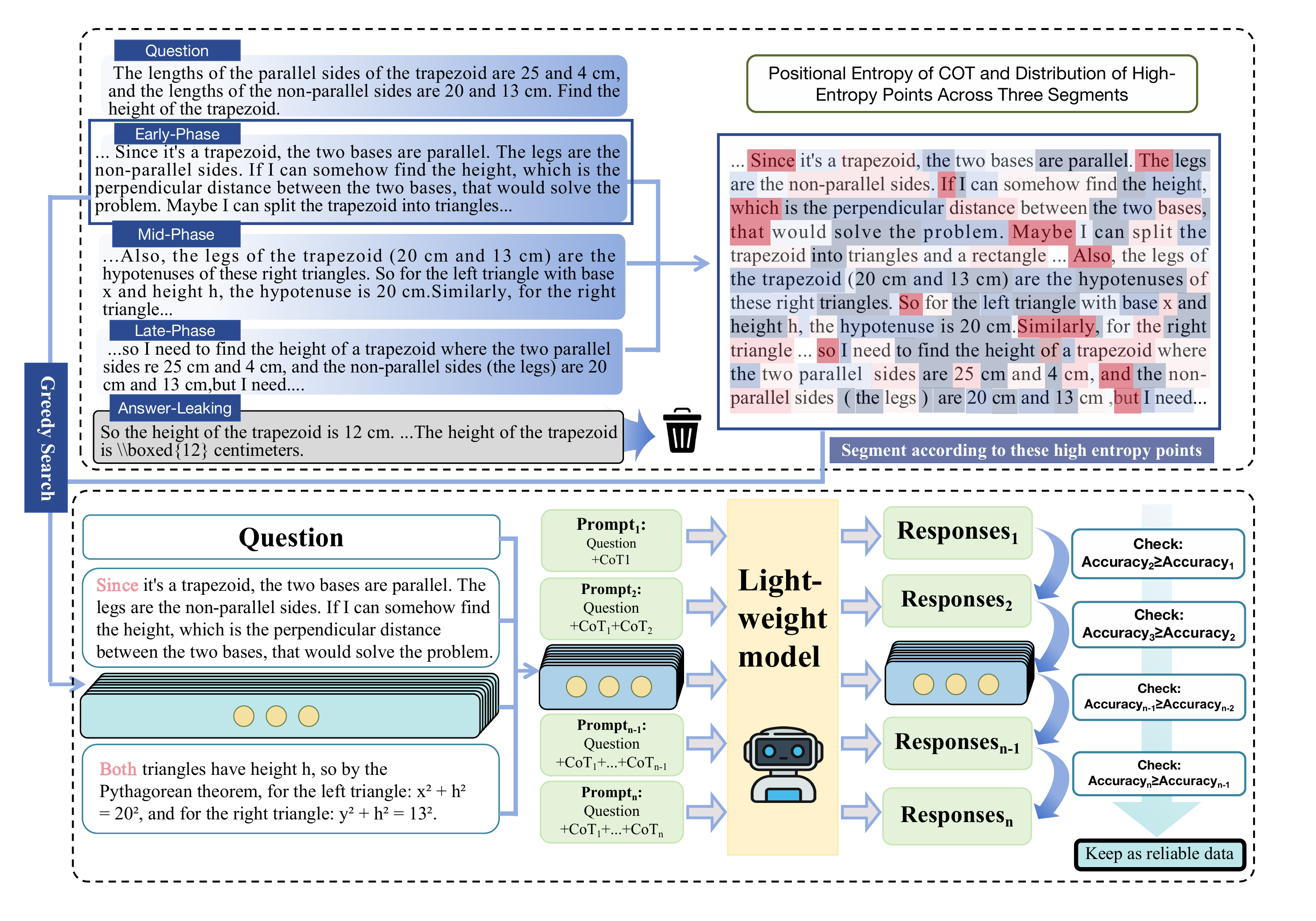} 
    \caption{Token-level entropy is computed over the CoT, and high-entropy positions are analyzed across early, middle, and late segments. Segmentation points are adaptively selected via greedy search to match the segment-wise entropy distribution, partitioning the CoT into sub-segments. The question combined with progressively accumulated CoT segments is used to prompt a lightweight model for multiple rollouts; the last segment is excluded to avoid answer leakage. Samples whose CoT segments' accuracy remains non-decreasing are retained as reliable data.}
\end{figure*}

For the $i$-th region, let $\mathcal{H}_i$ be the set of high-entropy positions, and let $s_i$ sub-segments be allocated to the region. The segmenting points are chosen greedily, starting from the smallest and largest high-entropy positions, and then iteratively selecting the position that maximizes the sum of distances to the current set of splitting points:
\[
h^\star = \arg\max_{h \in \mathcal{H}_i \setminus \mathcal{S}_i^{(m-1)}} \sum_{h' \in \mathcal{S}_i^{(m-1)}} |h - h'|.
\]
This procedure continues until $|\mathcal{S}_i^{(m)}| = s_i - 1$. By using a simple greedy algorithm, this strategy prevents the excessive concentration of high-entropy split points in local regions.

The final sequence of cutting points across all regions is merged and sorted to obtain the segmented reasoning sequence:
\[
\mathcal{T} = (\mathcal{T}_1, \dots, \mathcal{T}_N).
\]
These segments are then used to construct prefix-conditioned prompts for the subsequent stage, where they undergo stability evaluation using Monte Carlo rollouts.

\begin{algorithm}[t]
\caption{Entropy-Guided Progressive CoT Filtering}
\label{alg:CoT_filtering}
\begin{algorithmic}[1]
\Require Dataset $\mathcal{D}$, teacher model $\mathcal{M}_T$, rollout model $\mathcal{M}_g$
\Ensure Reliable dataset $\mathcal{D}_{\mathrm{rel}}$, deferred dataset $\mathcal{D}_{\mathrm{def}}$

\ForAll{$(x,y,\mathcal{T}) \in \mathcal{D}$}
    \State Compute token-level entropy $\{H_i\}$ using $\mathcal{M}_T$
    \State Select top-$K$ high-entropy positions $\mathcal{H}$
    \State Divide $\mathcal{T}$ into early, middle, and late regions
    \State Allocate segment numbers proportional to region-wise entropy counts
    \State Greedily select dispersed split points and segment $\mathcal{T}$ into $(\mathcal{T}_1,\dots,\mathcal{T}_N)$

    \For{$k=1$ to $N-1$}
        \State Construct prefix prompt $\mathcal{P}_k=\text{Concat}(x,\mathcal{T}_1,\dots,\mathcal{T}_k)$
        \State Estimate correctness probability $\hat{a}_k$ via $R$ Monte Carlo rollouts of $\mathcal{M}_g$
    \EndFor

    \If{$\hat{a}_1 \le \hat{a}_2 \le \dots \le \hat{a}_{N-1}$}
        \State Add sample to $\mathcal{D}_{\mathrm{rel}}$
    \Else
        \State Add sample to $\mathcal{D}_{\mathrm{def}}$
    \EndIf
\EndFor
\end{algorithmic}
\end{algorithm}

\subsection{Monte Carlo Rollout-based Prefix Evaluation}
\label{subsec:rollout}

A key property of high-quality responses is that their reasoning prefixes should gradually reduce uncertainty and increase the probability of generating the correct answer. We quantitatively evaluate this property using Monte Carlo rollouts.

For the segmented CoT, we construct a series of prefix-conditioned prompts:
\[
\mathcal{P}_k = \text{Concat}(x, \mathcal{T}_1, \dots, \mathcal{T}_k), \quad k=1,\dots,N-1,
\]
removing the final reasoning segment to avoid potential leakage of the answer information.

For each prefix $\mathcal{P}_k$, we perform $R$ independent generations using a lightweight model $\mathcal{M}_g$ and estimate the probability of generating the correct answer through Monte Carlo:
\[
\hat{a}_k \approx \mathbb{P}_{\mathcal{M}_g}(y \mid \mathcal{P}_k) = \frac{1}{R} \sum_{r=1}^R \mathbb{I}\left(\mathcal{M}_g(\mathcal{P}_k^{(r)}) = y\right),
\]
where $\mathbb{I}$ is the indicator function.

If the estimated probability sequence satisfies:
\[
\hat{a}_1 \le \hat{a}_2 \le \dots \le \hat{a}_{N-1},
\]
it indicates that each additional CoT segment enhances the probability of the model reaching the correct answer. We classify such samples as \emph{reliable data}, and they are retained for distillation training. %In addition, we observe that for some \emph{unreliable} samples, the rollout accuracy for every prefix is zero. This is likely because the problem is too difficult and thus beyond the capabilities of the lightweight model, or there is an error in the ground truth labeled in the dataset for this sample. Therefore, we store such data separately.

% -----------------------Experiment---------------------Let’s verify step by step
\section{Experiments}
\label{sec:experiments}

\subsection{Datasets and Evaluation Metrics}
\label{subsec:evaluation}

We employ a diverse suite of benchmarks categorized by difficulty: (1) \textbf{GSM8K} ~\cite{cobbe2021training} is a dataset consisting of grade-school level math problems that require logical reasoning, testing a model’s ability to solve elementary-level math problems; (2) \textbf{MATH-500} ~\cite{lightman2023let} is a curated subset of 500 representative high-school math problems, serving to assess a model’s ability to handle more advanced mathematical reasoning;  (3) \textbf{Gaokao} ~\cite{zhang2023evaluating} is a benchmark that includes mathematics problems from the Chinese National College Entrance Examination, designed to evaluate high school graduates’ comprehensive mathematical proficiency. These three datasets serve as basic-level benchmarks. For advanced competition-level assessment, we use (4) \textbf{AMC23} (Mathematical Association of America, 2023), which comprises challenging problems from the 2023 American Mathematics Competitions; (5) \textbf{MathOdyssey} ~\cite{fang2025mathodyssey}, a dataset spanning high-school to early undergraduate difficulty that emphasizes multi-step derivations and conceptual understanding; and (6) \textbf{OlympiadBench} ~\cite{he2024olympiadbench}, a collection of olympiad-level problems with non-standard formats requiring creative problem-solving strategies. We use exact match accuracy as the primary metric for evaluating performance, determined by comparing the predicted final answer, enclosed by \verb|\boxed|, with the ground-truth answer. All evaluations are conducted using the vLLM inference backend under identical generation parameters: zero-shot prompting with greedy decoding. To ensure statistical stability, we report the average exact-match accuracy across five independent runs for each model-dataset pair. Further implementation details are provided in Appendix ~\ref{app:B2}.

\begin{figure}[t]
    \centering
    \includegraphics[width=0.5\textwidth]{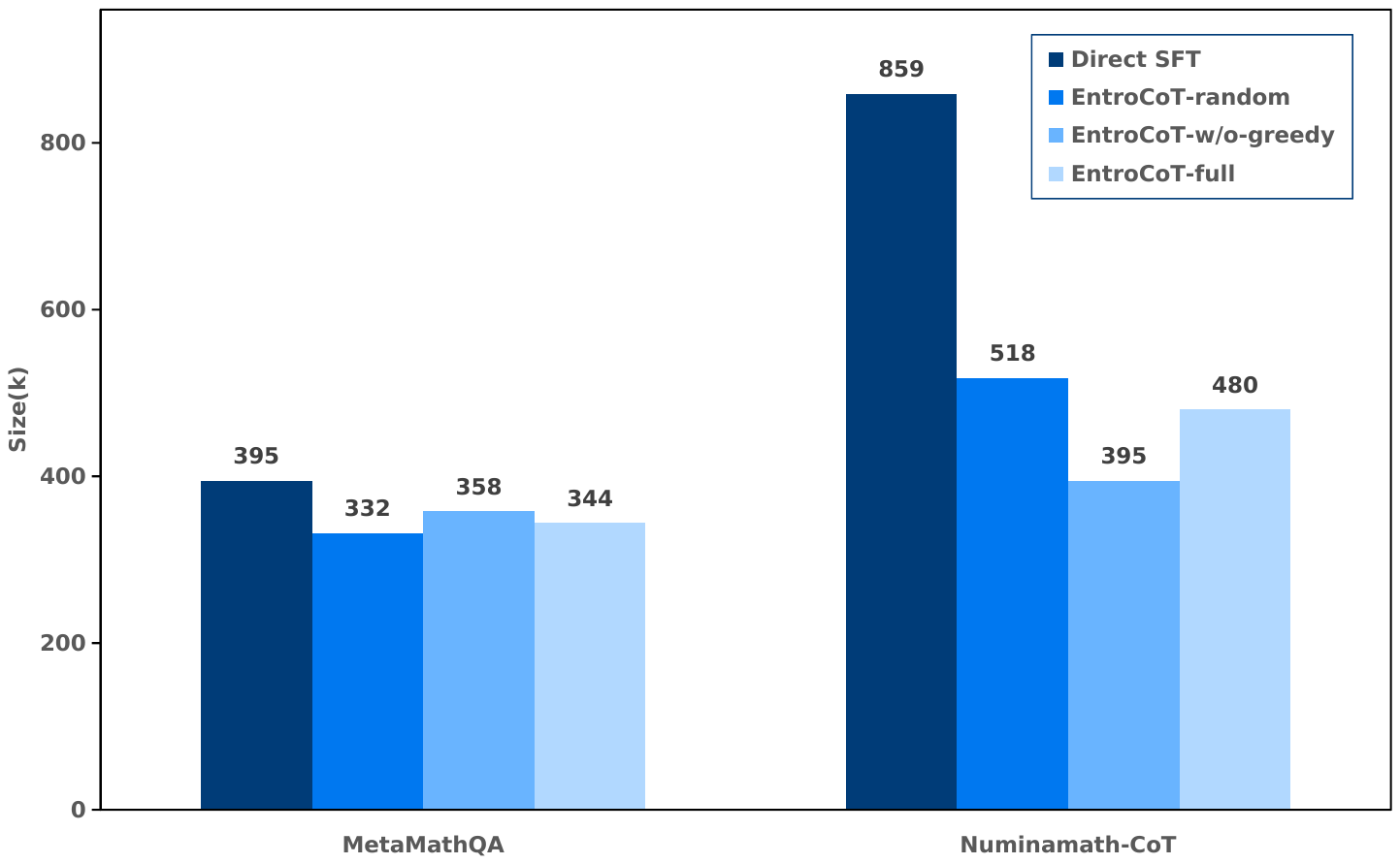}
    \caption{
        Dataset sizes for MetaMathQA and NuminaMath under different training strategies.
        Direct SFT denotes standard supervised fine-tuning, while EntroCoT variants
        differ in sampling and greedy constraints.
    }
    \label{fig:dataset_size}
\end{figure}

\begin{table*}[t]
    \centering
    \small
    \setlength{\tabcolsep}{3pt}
    \begin{tabularx}{\textwidth}{l r X ccc ccc c}
    \toprule
    \multicolumn{1}{c}{\textbf{Dataset}} &
    \multicolumn{1}{c}{\textbf{Size}} &
    \textbf{Method} &
    \multicolumn{3}{c}{\textbf{Basic Level}} &
    \multicolumn{3}{c}{\textbf{Competition Level}} &
    \textbf{Avg.} \\
    \cmidrule(lr){4-6} \cmidrule(lr){7-9}
    & & &
    \textbf{GSM8K} & \textbf{MATH} & \textbf{GaoKao} &
    \textbf{Odyssey} & \textbf{Olympiad} & \textbf{AMC23} &
    \\
    \midrule
    \multicolumn{10}{c}{\textbf{Meta-Llama3.1-8B}} \\
    \midrule

                   & 395k & Direct SFT
                   & \textbf{77.03} & 33.80 & 23.64
                   & 7.46 & 6.22 & 7.50 & 25.94 \\  

    MetaMathQA     & 332k & EntroCoT-random
                   & 74.83 & 31.40 & 23.90
                   & 6.68 & 5.93 & 5.00 & 24.62 \\

                   & 358k & EntroCoT-w/o-greedy
                   & 75.01 & 32.60 & 23.22
                   & 6.78 & \textbf{6.84} & 8.50 & 25.49 \\

                   & 344k & EntroCoT-full
                   & 76.89 & \textbf{35.80} & \textbf{27.01}
                   & \textbf{7.97} & 6.81 & \textbf{15.00} & \textbf{28.25} \\

    \midrule
                   & 859k & Direct SFT
                   & 72.10 & 37.20 & 32.73
                   & \textbf{20.82} & 13.04 & 19.00 & 32.48 \\

    NuminaMath     & 515k & EntroCoT-random
                   & 71.34 & 39.24 & 36.67
                   & 19.69 & 12.86 & 19.00 & 33.13 \\

                   & 395k & EntroCoT-w/o-greedy
                   & 70.96 & 39.80 & 38.96
                   & 17.48 & 12.00 & 17.50 & 32.78 \\

                   & 480k & EntroCoT-full
                   & \textbf{76.00} & \textbf{41.20} & \textbf{40.00}
                   & 19.54 & \textbf{14.37} & \textbf{20.00} & \textbf{35.19} \\

    \midrule
    \multicolumn{10}{c}{\textbf{Qwen2.5-Math-1.5B}} \\
    \midrule

                   & 395k & Direct SFT
                   & 48.60 & 33.84 & 33.72
                   & 17.12 & 10.28 & 7.50 & 25.18 \\

    MetaMathQA     & 332k & EntroCoT-random
                   & 45.40 & 33.92 & 35.58
                   & 16.45 & \textbf{12.12} & 8.50 & 25.33 \\

                    & 358k & EntroCoT-w/o-greedy
                   & 47.43 & 34.44 & 35.12
                   & 16.20 & 10.67 & 8.50 & 25.39 \\

                   & 344k & EntroCoT-full
                   & \textbf{50.19} & \textbf{34.56} & \textbf{37.35}
                   & \textbf{17.23} & 11.14 & \textbf{15.00} & \textbf{27.58} \\

    \midrule
                   & 859k & Direct SFT
                   & 70.90 & 54.64 & 46.07
                   & 21.44 & 19.73 & 32.50 & 40.88 \\

    NuminaMath     & 515k & EntroCoT-random
                   & 71.01 & 52.12 & 44.21
                   & 22.67 & 21.48 & 32.50 & 40.67 \\

                   & 395k & EntroCoT-w/o-greedy
                   & 73.09 & 48.20 & 40.52
                   & 20.31 & 18.07 & 35.00 & 39.20 \\

                   & 480k & EntroCoT-full
                   & \textbf{74.65} & \textbf{59.60} & \textbf{48.80}
                   & \textbf{23.40} & \textbf{24.35} & \textbf{45.50} & \textbf{46.05} \\

    \bottomrule
    \end{tabularx}
    \caption{Main results (\%) on mathematical benchmarks. MATH, GaoKao, Odyssey, and Olympiad correspond to the MATH500, GaoKao2023EN, MathOdyssey, and OlympiadBenchEN benchmarks, respectively. Bold marks the best score per dataset. Avg. is derived by calculating the average accuracy of the six benchmarks.
}
    \label{tab:main_results}
\end{table*}

\subsection{Baselines}
We evaluate our method by establishing a series of baselines for comparison. The most important baseline is Direct-SFT. In addition, to disentangle the contribution of every design decision in EntroCoT, we conduct the ablation experiments. Below we describe several baselines that progressively isolate (i) the impact of entropy-based segmentation and (ii) the importance of the greedy dispersion heuristic.
(1) \textbf{Direct-SFT}: The primary reference that fine-tunes on the complete MetaMathQA-395\,k or NuminaMath-859\,k dataset without any sample removal or rewriting. This replicates the conventional distillation pipeline where every trace is treated as gold supervision.
(2) \textbf{EntroCoT-random}: We preserve the entire segmental rollout pipeline (\S3.2) but replace entropy-guided segmentation with random cutting points. High-entropy tokens are still computed, yet they are ignored when constructing $(\mathcal{T}_1,\dots,\mathcal{T}_N)$. This ablation tests whether token-level uncertainty genuinely marks logical fault lines or merely acts as a spurious feature correlated with sequence position.
(3) \textbf{EntroCoT-w/o-greedy}: We retain the entropy computation and the early/middle/late ternary split, but inside each third we \textit{randomly} choose $s_i-1$ segmentation points instead of applying the greedy max-sum-of-distances rule. The resulting segments therefore still concentrate around high-entropy regions, yet may cluster adjacent high-uncertainty tokens. Comparing with EntroCoT-full quantifies the benefit of spatially \textit{dispersing} segmentation anchors so that the lightweight model receives CoT which contains relatively more complete reasoning steps.
(4) \textbf{EntroCoT-full}: Our complete method: entropy-guided segmentation with greedy dispersion, rollout-based verification, and training \textit{only} on the samples that satisfy the reliable criterion. 

\subsection{Implementation Details}
\label{subsec:setup}

We empirically evaluate the Entropy-Guided Progressive CoT Filtering (EntroCoT) framework as a data-centric procedure for enhancing mathematical reasoning. Our experiments aim to determine whether filtering unreliable reasoning traces can outperform full-dataset training and to identify the essential components of our framework.  To evaluate the generality and effectiveness of our proposed approach, we perform supervised fine-tuning (SFT) experiments using MetaMathQA~\cite{yu2023metamath} and NuminaMath-CoT~\cite{li2024numinamath} datasets on representative base models. Specifically, we select Llama-3.1-8B~\cite{touvron2023llama} as a large-scale model of llama series, and Qwen2.5-Math-1.5B~\cite{yang2024qwen2} as a model of qwen series for finetune. For each dataset, we apply our method to partition the samples into three disjoint buckets: a reliable set of behaviorally beneficial CoTs, a reject set of logically harmful traces, and an all-zero set of extremely difficult questions or harmful traces. Specifically, we use Qwen3-4B-Instruct ~\cite{yang2025qwen3} as the lightweight model for rollout and the number of rollouts per round is set to \textit{r = 8}. Moreover, the number of segments is set to \textit{k = 5}. After applying our method on datasets, we get a total of \textit{m = 480,313} reliable samples for Numinamath-CoT, while for MetaMathQA~\cite{yu2023metamath}, we get a total of \textit{n = 344,405} reliable samples. In addition, Deepseek-R1~\cite{guo2025deepseek} is selected for the entropy calculation. To ensure comparability, we maintain a unified experimental configuration across all SFT variants in the same set of experiments. The detailed training and rollout settings, including learning rates, training epochs and so on, are provided in Appendix~\ref{app:training}.

\begin{comment}
All conditions are trained with the same recipe detailed in Appendix~D.1: AdamW, peak LR $1\!\times\!10^{-5}$, cosine decay with $10\%$ warm-up, global batch size $128$, and $2$--$3$ epochs on $8\!\times\!$NVIDIA~H200. Decoding at test time uses greedy generation (temperature $= 0$, top-p $= 1$) through vLLM; exact-match accuracy is extracted by \texttt{math-verify} and averaged over five random seeds on the six benchmarks of \S4.2. By contrasting EntroCoT with this graded set of baselines we can attribute empirical gains to the specific combination of entropy-guided localisation, greedy dispersion, and rollout-verified quality gating that constitutes our final system.
\end{comment}

\subsection{Main Results}
\label{subsec:main_results}
Table 1 presents our main experimental results across all benchmarks, base models, and training datasets. We observe several consistent patterns that highlight the effectiveness of our Entropy-Guided Progressive CoT Filtering (EntroCoT) approach.

\textbf{Filtering via EntroCoT consistently outperforms full-dataset training.}
Despite discarding 45\% of NuminaMath-CoT and 13\% of MetaMathQA, the “EPCoT-full” models surpass the Direct-SFT baseline on almost all benchmarks. The largest absolute gains appear on the most challenging splits: Llama-3.1-8B improves +7.50\% on AMC23 when training on MetaMathQA; Qwen2.5-Math-1.5B gains +13.00\% on AMC23 and +4.62\% on Olympaid. Averaged across all six datasets, EntroCoT yields +2.31\% for Llama-3.1-8B and +2.40\% for Qwen2.5-Math-1.5B when training on MetaMathQA and yields +2.71\% for Llama-3.1-8B and +5.17\% for Qwen2.5-Math-1.5B when training on Numinamath-CoT, mirroring the prior observation that higher-quality supervision disproportionately benefits competition-level reasoning. The outsized gains on long-chain problems stem from two factors. First, competition-grade questions require significantly more intermediate steps, so a single “early-error, late-correction” sequence constitutes an extended source of incorrect gradient updates; EntroCoT’s monotonicity test rejects such entire chains, removing the longest and most frequent negative signals. Second, the lightweight rollout verifier is already near-chance on these difficult items, so any contaminated prefix causes a sharp drop in estimated correctness, ensuring that almost all flawed long derivations are filtered out. Consequently, the retained set is enriched for clean, multi-step symbolic derivations, and the student model observes a higher proportion of valid long-range reasoning patterns during SFT, which directly translates into larger accuracy improvements on the hardest benchmarks.

\textbf{Entropy-guided segmentation is the key driver.}
EPCoT-random incurs consistent and non-negligible degradations across almost all benchmarks: on the Llama3.1-8B it drops accuracy by 2.06\% on GSM8K, 4.40\% on MATH-500, 3.11\% on GaoKao2023EN, and margins of 1.29\%, 0.88\% and 10.00\% on the competition-level Odyssey, OlympiadBenchEN and AMC23 benchmarks, respectively on MetaMathQA; similar trends appear for Qwen2.5-Math-1.5B, where average losses reach 5.24\% on basic tasks and 5.53\% on competition problems when training on Numinamath. These systematic deficits, summarised in an overall $-5.38\%$ average, fall below the EPCoT-full and demonstrate that random segmentation is actively harmful rather than merely ineffective. The root cause is the inability of random cuts to disentangle erroneous segments from subsequent correct ones: cut points frequently land within mixed-quality spans, producing prefixes that begin with a hallucinated claim but end with a valid derivation. The lightweight rollout model can still exploit the correct tail signal to reach the right answer, so estimated accuracy does not decline and the monotonicity criterion fails to reject the flawed prefix. As a result, early-stage mistakes survive filtering, are included in training, and measurably degrade generalisation performance.

\textbf{Greedy spatial dispersion is indispensable.}
EPCoT-w/o-greedy exhibits consistent performance degradation across all test suites. On MetaMathQA, the average accuracy of Llama-3.1-8B drops from 28.25\% to 25.49\%, while Qwen2.5-Math-1.5B falls from 27.58\% to 25.39\%. When training on the larger NuminaMath-CoT dataset, the two models exhibit further decrements of 2.41\% and 6.85\%, respectively, demonstrating that the performance degradation caused by removing greedy dispersion becomes more pronounced as the dataset size increases. This deterioration is attributable to the clustering of high-entropy cut points within a narrow 50-token window, which reduces the aggregate token coverage of adjacent segments and limits the behavioural diversity presented to the rollout verifier. Consequently, the estimated accuracy curves exhibit higher variance, and harmful reasoning fragments are less likely to be detected and removed. The greedy max-sum-of-distance rule mitigates this effect by maximising inter-cut spacing, thereby ensuring that each segment contains a rich and non-redundant portion of the reasoning chain and preserving the discriminative power of the filter.

\section{Discussion}
\textbf{Filtering generalises beyond mathematics.}
Although we benchmark on math, the pipeline makes no domain-specific assumptions. 
 When the same entropy-cut/rollout strategy is applied, it is also applicable to problems with rich reasoning processes such as synthetic chemical mechanism problems and physical problems. The key requirement is a verifiable final answer; domains such as open-ended creative writing or legal argumentation would need an external value function instead of exact-match grading, but the entropy signature itself remains informative.

\textbf{Rejected samples can be recovered.}
For samples that fail the monotonicity check, they can not be discarded immediately, but can be attempted to recover their inference traces. Let the first position where rollout accuracy $\hat{a}_k > \hat{a}_{k+1}$ occur at index $k^\star$. We regenerate candidate continuations starting from this segment using a stronger model $\mathcal{M}_R$ (e.g., GPT series), and only keep the candidates where the final answer is correct. The prefix evaluation process from Section~\ref{subsec:rollout} is then repeated for the recovered samples.
If any candidate satisfies the monotonicity condition, the repaired sample is added to the reliable dataset; otherwise, the sample is placed into a deferred dataset for more complex recovery strategies or manual analysis. %The code related is also provided in "software" appendix.

\textbf{Recovered-rejected pairs can be served as DPO treasure trove.}
The traces that are filtered out—along with their refined counterparts—naturally form paired preference data: the original (high-entropy, misleading) chain serves as the ``rejected'' response, while the entropy-guided repaired version becomes the ``chosen'' one.  
Feeding these pairs directly into Direct Preference Optimization (DPO) without any relabelling can enhance the model's capabilities at the reasoning step level, effectively recycling waste compute into an extra alignment signal.  
This observation suggests that future pipelines should \emph{store} rejected traces rather than delete them, turning the expensive filtering stage into a dual-purpose generator of both clean SFT data and cheap preference data for RLHF-style training.

\section{Conclusion}
We present \textbf{EntroCoT}, an entropy-guided pipeline that automatically filters ``answer-right-but-reasoning-wrong'' traces from large-scale CoT datasets. By segmenting each reasoning chain at its highest-uncertainty tokens and validating every segment with lightweight Monte-Carlo rollouts, we retain only samples whose intermediate steps monotonically increase the probability of the correct answer. Across six mathematical benchmarks and two base models, training on this reliable subset systematically outperforms full-dataset supervision, such as yielding average gains of \textbf{+2.71\,\%} on Llama-3.1-8B and \textbf{+5.17\,\%} on Qwen2.5-Math-1.5B while reducing training compute by up to \textbf{45\,\%} on Numinamath. The entropy-based spatial dispersion heuristic is shown to be indispensable: ablating it collapses verifier diversity and erases the entire accuracy advantage. These results demonstrate that \emph{reasoning quality}, not data quantity, is the decisive factor for eliciting robust mathematical generalization in distilled LLMs. EntroCoT can be dropped into any existing CoT fine-tuning workflow without architectural changes, offering a scalable path toward models that not only answer correctly but also reason correctly.

\section*{Limitations}
First, the overall pipeline is computationally expensive: for every candidate reasoning path we first perform a full forward pass with a strong model to calculate token-level entropy; subsequently, each trace is split into five segments and evaluated with eight independent Monte-Carlo rollouts(8 per segment × 5 segments by default).\\  
Second, %the refine-and-recover stage remains inefficient; when a sample fails the monotonicity check we currently fall back to a separate, stronger model (e.g., GPT-series) to regenerate continuations, incurring additional GPU hours and serial latency that scale linearly with the size of the “unreliable” bucket.  Developing a lighter, high-efficiency mechanism to repair unreliable reasoning paths in a single forward pass is left for future work.
EntroCoT’s rollout verifier relies on an exact final answer to estimate segment-wise accuracy; it therefore cannot handle proof verification, creative writing, legal argument, or any task where the “correct” conclusion is subjective or undefined. Extending the framework to tasks with only external, human-graded or debate-based validation signals is left to future work.
\bibliography{custom}

\clearpage
\newpage
\appendix
\section{Training and Rollout Details}
\label{app:training}
\begin{comment}
    
\subsection{Base Models and Data Preparation}

We consider two representative base models in our experiments:
(i) a general-purpose instruction-tuned LLM of approximately $8$B parameters
(\textbf{Model A}); and
(ii) a math-specialized LLM of comparable size (\textbf{Model B}).
Both checkpoints are standard open-weight models that have previously been
used for CoT-based mathematical reasoning.

For supervised fine-tuning (SFT), we adopt the same math instruction corpora
as in prior work~\cite{xu2025mind,zhao2025promptCoT}, in particular
\textbf{MetaMathQA} and \textbf{NuminaMath}-style CoT datasets.
Each example is stored as a question–answer pair, where the answer may
optionally contain an explicit chain-of-thought solution.
Before training, we tokenize all data with the native tokenizer of each base
model, and filter out examples that exceed the maximum context length after
concatenating the question and target sequence.

Our Entropy-guided Prompt Filtering (EntroCoT) procedure is applied offline on
the training corpora to obtain three disjoint buckets:
a \emph{realiable} (reliable) subset used for the main SFT runs, and two
diagnostic subsets containing clearly harmful and behaviorally useless
examples.
The EntroCoT algorithm itself is fully described in Section~\ref{sec:sca};
here we only specify the training configuration used once the filtered
subsets are produced.
\end{comment}

\subsection{Training Details}

We utilize Pai-Megatron as the SFT training framework. The initial learning rate is set to $1 \times 10^{-5}$ with a warm-up ratio of 0.1, and cosine scheduling is applied to gradually decay the learning rate to $1 \times 10^{-6}$. We apply sequence packing for training, while the sequence length is set to 32768 tokens, with a global batch size of 64 and a micro batch size as 1. For the parallel strategy, when training Llama-3.1-8B, we set tensor-parallel as 4 and pipeline-parallel as 2, and for Qwen2.5-Math-1.5B, we set both tensor-parallel and pipeline-parallel as 1. Llama-3.1-8B is trained for 3 epochs on MetaMathQA and NuminaMath-CoT while Qwen2.5-Math-1.5B is trained for 2 epochs on MetaMathQA and NuminaMath-CoT. All SFT experiments are conducted on \textbf{8$\times$~NVIDIA H20 GPUs}. For the training template, we adopt the following default style template.

\begin{promptbox}{Training Template}
\small\sffamily
\textbf{System:} You are a helpful AI assistant, who always ready to help user.\\[0.8em]
\textbf{User:}\\
\texttt{<question text>}\\[0.8em]
\textbf{Assistant:}\\
\texttt{<answer text>}\\
\end{promptbox}

\subsection{Rollout Details}
We use Qwen3-4B-Instruct as the lightweight model for rollout. In detail, the max new tokens is set as 8192 while top-p and top-k are set as 0.8 and 20 respectively. In addition, repetition penalty is set as 1.1. To accelerate, we use  \textbf{96$\times$~NVIDIA H20 GPUs} to perform Monte Carlo rollout in parallel. Moreover, we use sglang as the inference engine.

\subsection{Entropy Calculation Details}
We utilize DeepSeek-R1 as the model for entropy calculation. Specifically, four replication of DeepSeek-R1 is deployed on \textbf{64$\times$~NVIDIA H20 GPUs} while each of them is deployed on \textbf{16$\times$~NVIDIA H20 GPUs}. In addition, the tensor-parallel, expert-parallel and data parallel are set as 16, 16 and 8 respectively. Moreover, the parameter of logprobs is set as 5 to derive the top-5 highest values of log probabilities for entropy calculation.

\section{Evaluation}
\label{app:evaluation}

\begin{comment}
    
\subsection{Mathematics Benchmarks}
\label{app:math_benchmarks}

We evaluate all models on the same suite of mathematical reasoning
benchmarks as recent CoT tuning work~\cite{xu2025mind}, covering both
grade-school and competition-level tasks:

\begin{itemize}
    \item \textbf{GSM8K}:
    1{,}319 grade-school math word problems assessing multi-step arithmetic
    reasoning in natural language.
    \item \textbf{MATH500}:
    a curated subset of 500 representative problems from the MATH dataset,
    spanning diverse topics and difficulty levels.
    \item \textbf{Gaokao2023EN}:
    385 English-translated problems from the 2023 Chinese college entrance
    exam (Gaokao) mathematics section, testing advanced high-school problem
    solving.
    \item \textbf{MathOdyssey}:
    387 problems ranging from high school to early undergraduate difficulty,
    emphasizing multi-step derivations and conceptual understanding.
    \item \textbf{OlympiadBenchEN}:
    675 olympiad-level problems with non-standard formats and high demands
    on creative reasoning.
    \item \textbf{AMC23}:
    40 problems from the 2023 American Mathematics Competition (AMC),
    providing a standardized competition setting.
\end{itemize}

Following~\citet{xu2025mind}, we group benchmarks into a
\emph{basic-level} split (e.g., GSM8K) and a \emph{competition-level} split
(e.g., AMC23, OlympiadBenchEN), and report exact-match accuracy on the final
answer for each dataset, as well as averages over the basic, competition,
and overall sets.
\end{comment}

\subsection{Evaluation Prompt}
\label{app:eval_prompts}
We extract final answers by extracting the content included in \verb|\boxed{}| and applying normalization rules (e.g., trimming spaces, canonicalizing fractions).
Each problem is wrapped in a template that explicitly encourages step-by-step reasoning and the template is referenced to the selected benchmark. For example, the template for AMC23 is as following:

\begin{promptbox}{Prompt Template}
    \small\sffamily
    You are an expert mathematician specializing in competition-level mathematics. Solve the given AIME problem step by step and provide your final answer.\\[0.5em]
    
    \textbf{Guidelines:} 
        \begin{itemize}
        \item AIME answers are always integers from 0 to 999
        \item Show your complete reasoning process
        \item Provide the final answer as an integer enclosed in \verb|\boxed{}|
        \item Be precise and rigorous in your mathematical reasoning
    \end{itemize} 
    
    \textbf{Problem:} \\
    $<$problem$>$ \\[0.5em]
    
Please solve this problem step by step. Remember that the answer must be an integer from 0 to 999. Provide your final answer in \verb|\boxed{}| format.

\end{promptbox}

This evaluation pipeline is kept identical across all methods to ensure a
fair comparison of different strategies.

\subsection{Evaluation Settings}
\label{app:B2}
\label{subsec:eval_settings}

The maximum token limit for generation was set to 4096. It is noteworthy that, although we set the decoding temperature parameter to zero, vLLM(inference engine we use for evaluation) outputs still exhibited some randomness. For answer extraction and comparison, we employed the Math-Verify tool. Given limitations of math-verify in handling complex expressions, responses failing initial verification were subsequently validated using DeepSeek-R1. All model evaluations were performed using \textbf{1$\times$~NVIDIA H20 GPU}.

\section{Case Study}
\label{app:case_study}

\subsection{Reasoning and Annotation Errors}

Cases of the samples containing reasoning or annotation
errors are provided. In particular, some errors arise from arithmetic mistakes or
semantic misinterpretations, while others stem from the inappropriate
selection of problem-solving strategies. In the following case studies,
we explicitly identify the \emph{error onset} and analyze how the
subsequent reasoning is affected.

\begin{promptbox}{Case Study 1: Zinc Intake Problem}
\small\sffamily

\textbf{Problem:}\\
Jerry takes two antacids that weigh 2 grams each and contain 5\% zinc by weight.
He then takes $x$ smaller antacids that weigh 1 gram each and contain 15\% zinc.
The total amount of zinc consumed is 650 milligrams. What is the value of $x$?

\vspace{0.6em}
\textbf{Answer from Dataset:} \\[0.3em]
To solve this problem, we need to determine the value of $x$, which represents the number of smaller antacids that Jerry takes.

Let's break down the information given:

Weight of each larger antacid: 2 grams

Zinc content of each larger antacid: 5\%

Weight of each smaller antacid: 1 gram

Zinc content of each smaller antacid: 15\%

Total amount of zinc consumed: 650 milligrams

To find the amount of zinc from the larger antacids, we multiply the weight of each antacid by the zinc content:

\errorbox{2 grams $\times$ 5\% = 0.1 grams of zinc}

To find the amount of zinc from the smaller antacids, we multiply the weight of each antacid by the zinc content:

1 gram $\times$ 15\% = 0.15 grams of zinc

Now, we can set up the equation based on the total amount of zinc consumed:

0.1 grams $+ x \times 0.15$ grams = 650 milligrams

Let's convert the units to grams:

0.1 $+ 0.15x = 0.65$ grams\\Let's solve for x:\\0.15x = 0.65 - 0.1\\0.15x = 0.55\\To isolate x, we divide both sides of the equation by 0.15:\\x = 0.55 / 0.15\\x = 3.67\\The value of x is approximately 3.67. Since we cannot have a fraction of an antacid, we can conclude that Jerry takes 3 smaller antacids. The value of x is 3.

\vspace{0.6em}
\textbf{Error Analysis:}\\
The error occurs in the computation of zinc contributed by the larger antacids.
While $2 \text{ g} \times 5\% = 0.1 \text{ g}$ is correct \emph{per antacid},
the dataset fails to account for the fact that Jerry consumes two such antacids.

The correct zinc contribution from the larger antacids is therefore:
\[
2 \times 2 \text{ g} \times 5\% = 0.2 \text{ g}.
\]

After correction, the total zinc equation becomes:
\[
0.2 + 0.15x = 0.65.
\]

Solving yields:
\[
0.15x = 0.45 \quad \Rightarrow \quad x = 3.
\]

\vspace{0.3em}

\datasource{\textsc{MetaMathQA}}
\end{promptbox}

\begin{promptbox}{Case Study 2: Meeting Attendance Table}
\small\sffamily

\textbf{Problem:}\\
Each person marks the table with an X on the day(s) he or she would not be
able to attend a meeting. On which day can the most people attend the
meeting?

\vspace{0.6em}
\begin{center}
\begin{tabular}{|l|c|c|c|c|c|}
\hline
      & Mon & Tues & Wed & Thurs & Fri \\ \hline
Anna  & X   &      & X   &       &     \\ \hline
Bill  &     & X    &     & X     & X   \\ \hline
Carl  & X   & X    &     & X     & X   \\ \hline
\end{tabular}
\end{center}

\vspace{0.6em}
\textbf{Answer from Dataset:} \\[0.3em]
\\[0.6em]
\errorbox{
We can count the number of X's for each day to see which day has the most people unavailable.
}
\\[0.6em]
On Monday, there is 1 X. On Tuesday, there are 2 X's. On Wednesday, there are 0 X's. On Thursday, there are 2 X's. On Friday, there are 1 X. Therefore, the day with the most people available is \verb|\\boxed{Wednesday}|.The answer is: Wednesday.

\vspace{0.6em}
\textbf{Error Analysis:}\\
The error occurs at the very first reasoning step. The problem asks for the
day on which the most people \emph{can} attend the meeting. However, the
reasoning path instead focuses on counting the number of people who \emph{cannot}
attend.

This semantic inversion fundamentally alters the task objective. By
optimizing for maximum unavailability rather than maximum availability,
the sample's reasoning path proceeds in the wrong direction from the outset,
rendering the entire solution invalid regardless of any subsequent
calculations.

\datasource{\textsc{MetaMathQA}}
\end{promptbox}

\vspace{1em}

\begin{promptbox}{Case Study 3: Dune Buggy Average Speed}
\small\sffamily
\textbf{Problem:} \\
Conner rides a dune buggy in the desert. On flat sand, it travels at
60 miles per hour. On downhill slopes, it travels 12 miles per hour
faster than on flat sand. On uphill slopes, it travels 18 miles per hour
slower than on flat sand. If Conner rides his dune buggy one-third of the
time on flat sand, one-third of the time on uphill slopes, and one-third
of the time on downhill slopes, what is his average speed in miles per
hour? \\[0.6em]

\textbf{Answer from Dataset:} \\[0.3em]
To find the average speed, we need to consider the speeds at which Conner rides on each type of terrain and the time spent on each.

On flat sand, the dune buggy rides at 60 mph.

On downhill slopes, it rides at 60 mph + 12 mph = 72 mph.

On uphill slopes, it rides at 60 mph - 18 mph = 42 mph.
\\[0.6em]
\errorbox{Since Conner rides one-third of the time on each type of terrain, we can calculate the average speed by taking the harmonic mean of the three speeds.}

The harmonic mean is more appropriate than the arithmetic mean for average speed because it takes into account the time spent at each speed, rather than just the distances. The formula for the harmonic mean (H) of three numbers (a, b, c) is given by:\\$H = 3 / (1/a + 1/b + 1/c)$ \\Plugging in the values for a, b, and c:\\$H = 3 / (1/60 + 1/72 + 1/42)$\\First, find the common denominator for the fractions, which is the least common multiple (LCM) of 60, 72, and 42. The LCM of 60, 72, and 42 is 360.\\Now convert the fractions:
\\$1/60 = 6/360$
\\$1/72 = 5/360$
\\$1/42 = 8.57142857/360 (approximately)$
\\Now add the fractions:
$6/360 + 5/360 + 8.57142857/360 = 19.57142857/360$
Now take the reciprocal to find the harmonic mean:
\\$H = 3 / (19.57142857/360)$
\\$H = 3 * (360/19.57142857)$
\\$H \approx 3 * 18.39795918$
\\$H \approx 55.19387755$
\\Conner's average speed is approximately \verb|\\boxed{58}| miles per hour.\\

\vspace{0.6em}
\textbf{Error Analysis:} \\
The reasoning becomes incorrect at the boxed step. The problem specifies
that Conner spends an equal fraction of \emph{time} on each terrain type,
in which case the average speed should be computed using the
\emph{arithmetic mean}, not the harmonic mean. The harmonic mean is only
appropriate when equal \emph{distances} are traveled at different speeds.
This incorrect choice of averaging strategy leads to an erroneous final
answer.
Since Conner spends equal time on each terrain type (one-third each), the average speed is the arithmetic mean of the three speeds.

Given:
\begin{itemize}
    \item Flat sand: $60$ mph
    \item Downhill: $60 + 12 = 72$ mph  
    \item Uphill: $60 - 18 = 42$ mph
\end{itemize}

The average speed is:
\[
v_{\text{avg}} = \frac{60 + 72 + 42}{3} = \frac{174}{3} = 58 \text{ mph}
\]

\datasource{\textsc{NuminaMath-CoT}}
\end{promptbox}

\section{Use of AI}
We use LLM to help polish the sentences in the paper and correct grammatical errors.

\end{document}